# People detection and social distancing classification in smart cities for COVID-19 by using thermal images and deep learning algorithms

Abdussalam Elhanashi, Sergio Saponara, Alessio Gagliardi Dip. Ingegneria dell'Informazione, University of Pisa

*Abstract* - **COVID-19 is a disease caused by severe respiratory syndrome coronavirus. It was identified in December 2019 in Wuhan, China. It has resulted in an ongoing pandemic that caused infected cases including some deaths. Coronavirus is primarily spread between people during close contact. Motivating to this notion, this research proposes an artificial intelligence system for social distancing classification of persons by using thermal images. By exploiting YOLOv2 (you look at once), a deep learning detection technique is developed for detecting and tracking people in indoor and outdoor scenarios. An algorithm is also implemented for measuring and classifying the distance between persons and automatically check if social distancing rules are respected or not. Hence, this work aims at minimizing the spread of the COVID-19 virus by evaluating if and how persons comply with social distancing rules. The proposed approach is applied to images acquired through thermal cameras, to establish a complete AI system for people tracking, social distancing classification, and body temperature monitoring. The training phase is done with two datasets captured from different thermal cameras. Ground Truth Labeler app is used for labeling the persons in the images. The achieved results show that the proposed method is suitable for the creation of a smart surveillance system in smart cities for people detection, social distancing classification, and body temperature analysis.**

## I. Introduction

COVID-19 is a disease caused by new coronavirus which appeared in China in December 2019. COVID-19 symptoms include mainly fever, cough, chill, and shortness in breathing, body aches, loss of smell. COVID-19 can be severe, and in several cases it caused death. The coronavirus can spread from one person to another as diagnosed by researchers in laboratories. This pandemic spread to over 188 countries around the world [1] On June 30, 2020, WHO (world health organization) declared there have been 10.185.374 confirmed COVID-19 cases and 503.862 deaths [2]. There is no vaccine yet. Prevention involves wearing masks, washing-hands frequently and infected person should stay at home when people are sick to prevent spreading this pandemic to the others. This situation forces the global communities and governments to find the best mitigation plan to stop the spread of coronavirus. Social distancing has been claimed as the best solution to minimize the spread of this virus between people. It has been reported that all infected countries who applied the lock-down for their communities achieved a reduction of the number of COVID-19 cases and of the number of deaths from this pandemic.

Fever or chill is one of the common symptoms for coronavirus. Researchers in China found that 99 % of people infected with the coronavirus had a high temperature. Temperature screening is one method that can define the early symptoms for COVID-19 infection as declared by the world health organization.

This research aims at mitigating the spread of this virus and save the life of people and communities. In this work, we propose a deep learning model YOLOv2 for people detection in combination with an implemented algorithm for social distancing classification on thermal images.

Hereafter, the paper is organized as follows: after the introduction of COVID-19 in Section I, Section II presents a methodology for the proposed approach. Section III shows the experiment and results of this research. Conclusions and future work are drawn in Section IV.

## II. Methodology

*A. Neural Network Design*

A Deep Neural Network (DNN) application is used in MATLAB to construct the YOLOv2 neural network layers. Then the designed DNN is ported in embedded platforms like NVIDIA Jetson Nano. We built a CNN with 29 layers. This is to establish a light-weight model to fit the real-time implemnetaion of CNN inference also in low-cost embdded platfroms, such those of IoT nodes. The neural network layers include the input layer, middle layers, and YOLOv2 layers.

The proposed approach starts with the input image layer, which introduces the input image with size of (224 x 224 x 3) for our detector. Set of middle layers was used, which includes batch normalization, convolutional, max-pooling, and Relu (rectified linear unit) layers. Convolutional layers were used to map the features for the images. The size of filter was set to (3 x 3). It defines the height and width of the regions in the input image. Batch normalization layers were used to regularize the model and eliminate the overfitting problem. ReLU layers were utilized to introduce the non-linearity to the neural network. Maxpooling layers were used to downsample the images into pooling regions. We applied (2 x 2) for size of pooling with a stride of (2 x 2) for all max-pooling layers in a neural network.

'ReLU_5' was used as the feature extraction layer. This is to extract the features from neural network layers and then given as input to the YOLOv2 subnetwork layers. YOLO2 Layers were used in this detector which constructs the YOLOv2 detection network. The transform layer was utilized in YOLOv2 detector to stabilize the network for object localization. Finally, YOLOv2 output layer was used which refines the location of bounding boxes to the detected objects. The model was examined with a neural network analyzer and reported zero errors.

*B. Algorithm for Distancing Classification*

We also implemented code in MATLAB to work with bounding boxes of a detected person in the thermal images. This code classifies and decides if persons in the image are within safe distancing or not. We assigned a green color for safe social distancing and red color for unsafe social distancing for bounding boxes. First we find the number of persons in the images. If it is one person, green color is assigned for a bounding box of detected persons. When we have 2 or more persons, then color is decided from the function which is called findColor. This function will determine if the bounding box is 2 or more and in addition to that, it will calculate the distance between the centers of bounding boxes. The distance between the center of bounding boxes is calculated by using Euclidean formula, see Eq (1), where the distance between pixels is translated in a metric distance (knowing the range and field of view covered by the camera) and then compared to a threshold value. In case find-color function detects two bounding boxes and the distance is less than the threshold value, these boxes will have a red color. And if this function detects two bounding boxes and the distance is more than the threshold value, the color will be green for these boxes.

$$D = \sqrt{(x_2 - x_1)^2 + (y_2 - y_1)^2} \quad (1)$$

where D is the distance between the centers of bounding boxes

### III. EXPERIMENT AND RESULTS

We train the designed YOLOv2 with two different datasets of thermal images. The Dataset I consists of 553 images. These images were collected from different sources on the internet. The Dataset_II consists of 428 images. These images are infrared images that were created by FLIR company for thermal cameras [3]. We used ground truth labeler application in MATLAB for labeling the persons in the thermal images [4]. We split the images into (70% for training, 20% for validation, and 10% for testing) for each dataset. The model has been trained with stochastic gradient descent (sdgm) [5]. We fine-tuned the learning rate for training process at $10^{-3}$ to control the model change in response to the error. The mini-batch loss was recorded at 0.2. The trained model was run with an implemented algorithm in MATLAB script. Based on the results from testing Dataset, the proposed method achieved good performance for people detection, social and unsocial distancing classification on thermal images in both datasets, see Fig 1. We Evaluate precision metrics for person detection on the testing Dataset. We recorded average precision at 0.9 and miss-rate at 0.1 see Fig 2.

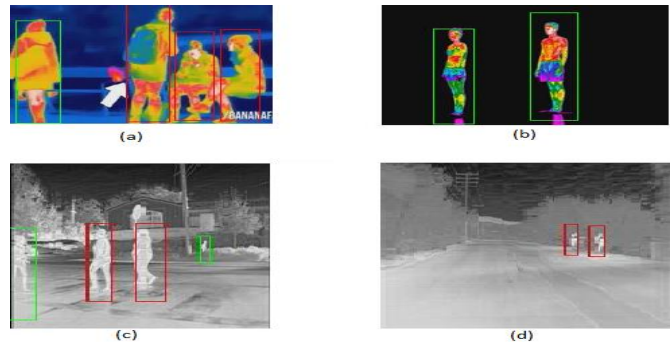

Fig 1: Sample Images from (a,b) Dataset_I, (c,d) Dataset_II

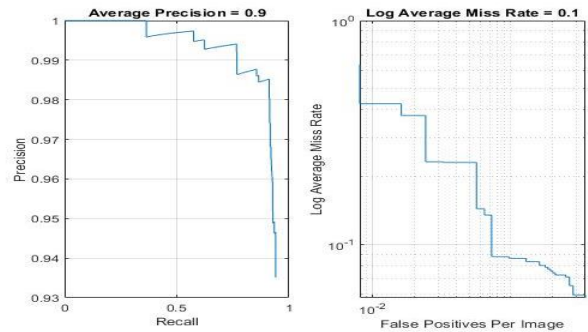

Fig 2: Results obtained by the proposed system in terms of Average precision & Average miss-rate

### IV. CONCLUSION AND FUTURE WORK

This research presented an intelligent surveillance system for people tracking and social distancing classification based on thermal images. YOLOv2 showed accurate results for people detection in terms of evaluating the average precision and average miss rate metric. A specific algorithm was implemented on bounding boxes to distinguish between safe and unsafe conditions, respectively marking as green and red the bounding box in the output image. In the future, we will extend our research to use video-camera to stream real state for people detection, social distancing, and temperature analysis for people. We will also deploy algorithm to low-cost IoT devices such as NVIDIA Jetson Nano. In addition to that, we will utilize this methodology on mobile cameras, e.g. mounted on an autonomous drone system, and hence drones are simpler to operate and more effective to capture fast actions of the detected objects from different angles.